\documentclass{article}
\usepackage[square,sort,comma,numbers]{natbib}
\usepackage{arxiv}
\usepackage{nicefrac}       
\usepackage{microtype}      
\usepackage{booktabs}       
\usepackage{natbib}
\usepackage{doi}
\usepackage{adjustbox}
\usepackage{amsmath}
\usepackage{amssymb}
\usepackage{caption}
\usepackage{float}
\usepackage[T1]{fontenc}
\usepackage{graphicx}
\usepackage{hhline}
\usepackage[utf8]{inputenc}
\usepackage[english]{babel}
\usepackage{mathtools}
\usepackage{multicol}
\usepackage{multirow}
\usepackage{subcaption}
\usepackage{wasysym}
\usepackage{hyperref}

\title{Image-based Method for Measuring and Classification of iron ore pellets using Star-Convex Polygons}

\author{Artem Solomko, Oleg Kartashev, Andrey Golov, Mikhail Deulin, Vadim Valynkin, Vasily Kharin
}

\renewcommand{\headeright}{\footnotesize {}}
\renewcommand{\undertitle}{\footnotesize {}}
\renewcommand{\shorttitle}{Image-based Method for Measuring and Classification of iron ore pellet using Star-Convex Polygons}

\hypersetup{
pdftitle={Image-based Method for Measuring and Classification of iron ore pellets using Star-Convex Polygons},
pdfsubject={q-bio.NC, q-bio.QM}, 
pdfauthor={},
pdfkeywords={},
}

\begin{document}
\maketitle

\begin{abstract}
\textit{We would like to present a comprehensive study on the classification of iron ore pellets, aimed at identifying quality violations in the final product, alongside the development of an innovative image-based measurement method utilizing the StarDist algorithm, which is primarily employed in the medical field. This initiative is motivated by the necessity to accurately identify and analyze objects within densely packed and unstable environments. The process involves segmenting these objects, determining their contours, classifying them, and measuring their physical dimensions. This is crucial because the size distribution and classification of pellets — such as distinguishing between ``nice'' (quality) and ``joint'' (caused by the presence of moisture or indicating a process of production failure) types — are among the most significant characteristics that define the quality of the final product.
Traditional algorithms, including image classification techniques using Vision Transformer (ViT), instance segmentation methods like Mask R-CNN, and various anomaly segmentation algorithms, have not yielded satisfactory results in this context. Consequently, we explored methodologies from related fields to enhance our approach. The outcome of our research is a novel method designed to detect objects with smoothed boundaries. This advancement significantly improves the accuracy of physical dimension measurements and facilitates a more precise analysis of size distribution among the iron ore pellets. By leveraging the strengths of the StarDist algorithm, we aim to provide a robust solution that addresses the challenges posed by the complex nature of pellet classification and measurement.}
\end{abstract}


\section{Introduction}

Iron ore pellets (\textit{further, pellets}) are a semi-finished product of metallurgical iron production. They are produced from the enrichment of iron-containing ores through pelletization and sintering. The agglomerate is the main component of the iron-containing part of them used in blast furnace production for the production of cast iron.

The process of making pellets is often called pelletizing. The charge, in the form of a mixture of finely ground iron-containing concentrates, fluxes (additives that regulate the composition of the product), and strengthening additives (\textit{usually}, bentonite clay), is moistened and subjected to rounding in rotating bowls or pelletizing drums.

As a result of pelletizing in special units —granulators— close to spherical particles with a diameter of 10-20 mm are obtained. To determine how well the process went, you need to know the dimensions and types, e.g., stones (leftovers from the pelletizing drums), joint pellets, nice or bad pellets. Analysis of the qualitative composition is carried out using laboratory experiments, which can be time-consuming and require specialized equipment. However, with the advancement of computer vision and machine learning technologies, it is now possible to develop more efficient and cost-effective methods for analyzing pellet quality.

We propose a method based on a supervised deep learning algorithm using a 2D camera, which can capture high-resolution images of the pellets. The algorithm is trained on a dataset of labeled images, where each image is annotated with the pellets' type. Once trained, the algorithm can accurately classify new images of pellets into different categories, such as stones, joint pellets, nice, or bad pellets. This approach has the potential to significantly reduce the time and cost associated with laboratory experiments, while also improving the accuracy and consistency of pellet quality analysis. Additionally, the use of a 2D camera and deep learning algorithm can also provide other benefits, such as:

\begin{itemize}
	\item \textit{Increased speed}: The algorithm can analyze images of pellets in real-time, allowing for faster quality control and decision-making.  

	\item \textit{Improved accuracy}: The algorithm can detect subtle variations in pellet quality that may not be visible to the human eye.

\end{itemize}
Overall, our proposed method has the potential to revolutionize the field of pellet quality analysis, enabling faster, more accurate, and more cost-effective analysis of pellet quality.

The main problem in detecting densely located objects (Fig. 1) is that identification and instance segmentation are difficult for those cases. Separating one object from another to determine their sizes and classes is quite challenging, and in tasks requiring high precision, it is extremely critical for understanding the granulometric composition of the current process.

        \begin{figure}[!htbp]
\centering
\begin{subfigure}[b]{0.45\textwidth}
\centering
\includegraphics[width=\textwidth]{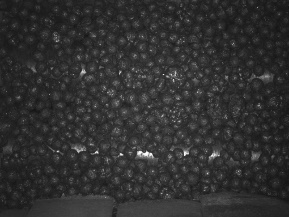}
\end{subfigure}
\hfill
      \begin{subfigure}[b]{0.45\textwidth}
\centering
\includegraphics[width=\textwidth]{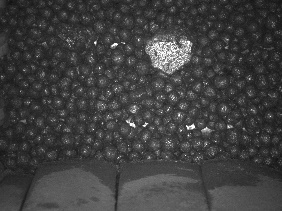}
\end{subfigure}
\end{figure}

                       \begin{figure}[!htbp]
\centering
\begin{subfigure}[b]{0.45\textwidth}
\centering
\includegraphics[width=\textwidth]{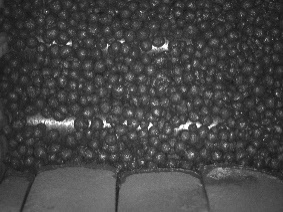}
\end{subfigure}
\hfill
     \begin{subfigure}[b]{0.45\textwidth}
\centering
\includegraphics[width=\textwidth]{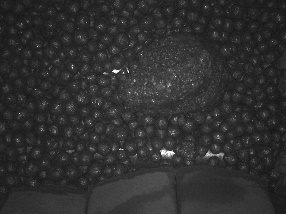}
\end{subfigure}
\end{figure}

\begin{center}
	{\scriptsize Fig. 1. The 2D-camera's region of interest is represented in the image – densely spaced objects make it difficult to separate them into classes.}
\end{center}

It became clear that it was necessary to look for a non-standard solution to the problem. The focus shifted to the medical domain, which places high demands on the accuracy of algorithms capable of working with densely located objects. In the medical field, computer vision and machine learning algorithms are widely used for tasks such as tumor detection, organ segmentation, and cell tracking. These algorithms are designed to handle complex and nuanced visual data and are often trained on large datasets of medical images.

One of the key challenges in the medical domain is the need to accurately detect and analyze small, densely packed objects, such as cells or cell nuclei. This requires algorithms that can handle high levels of visual complexity and accurately distinguish between different types of objects.

By drawing inspiration from the medical domain, we realized that similar approaches could be applied to the problem of pellet quality analysis. By leveraging the advances in computer vision and machine learning made in the medical domain, we could develop algorithms capable of accurately detecting and analyzing small, densely packed objects, such as pellets.

Overall, by looking to the medical domain for inspiration \cite{weigert2022nuclei}, we were able to develop a novel solution to the problem of pellet quality analysis, one that leverages the advances in computer vision and machine learning made in this field \cite{graham2021}. 

The most similar domain seemed to be the detection of cellular structures, namely cell nuclei (Fig. 2). The StarDist algorithm \cite{schmidt2018cell} \cite{weigert2020star} is suitable for such problems, which formed the basis for the development of the solution.

\begin{figure}[!htbp]
\centering
\begin{subfigure}[b]{0.85\textwidth}
\centering
\includegraphics[width=\textwidth]{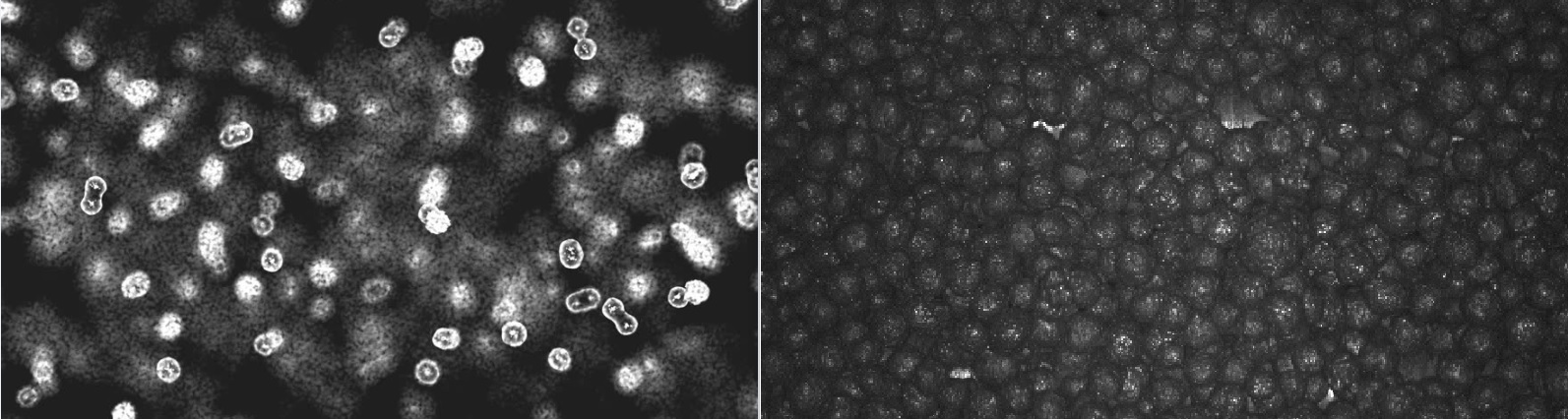}
\end{subfigure}

\begin{center}
	{\scriptsize Fig. 2. Image of a cells with nuclei (\textit{left}) and pellets (\textit{right}).}
\end{center}
\end{figure}
\section{Method}

We implemented the classic StarDist calculation method, which is a well-established approach in the field of object detection and segmentation. This method is distinctive because it allows us to predict the shapes of objects of interest directly, without the need for traditional bounding boxes. Instead, we utilize a contour-based representation known as the star-convex shape, which employs distance regression techniques to accurately map the boundaries of the objects.

In our application of this method, we also incorporate Non-Maximum Suppression (\textit{NMS}), a technique commonly used to eliminate redundant overlapping bounding boxes and retain only the most relevant predictions. However, in our case, \textit{NMS} is applied in a manner that focuses on refining the pixel-level segmentation for each specific object. This approach enables us to achieve results of notably high quality, as it enhances the precision of the object contours and minimizes errors in the segmentation process.

Furthermore, during the training phase of our model, we applied a technique known as \textit{pyramid patch weight smoothing} to the output. This technique involves adjusting the weights of the patches in a multi-scale manner, which helps to improve the robustness and accuracy of the predictions. By smoothing the weights, we can better capture the variations in object shapes and sizes, leading to more reliable segmentation results across different scales. After predicting the mask, we applied weight smoothing to the boundaries, which subsequently affected the final pixel classification. The mask was confirmed by morphological operations that expanded its boundaries, and this parameter was subsequently applied to the inference

The training was carried out by evaluating not only \textit{IoU} or other segmentation metrics, but also \textit{precision, recall}, and \textit{F1-score}, measured per pixel. It was important for us not only to accurately predict masks but also to accurately classify image pixels. Saving the best models was done based on the ugly class, because it is an indicator of production violations and shows dirt in the process.

Overall, the combination of these advanced methodologies contributes to the effectiveness of our object detection and segmentation efforts \cite{lefuvre2007}. More details are presented in Section 3.

\section{Experiments}

\subsection{Dataset}

At the outset of our project, we faced the critical task of determining the most effective method for creating annotations for our dataset. Upon initial evaluation, it became clear that the annotation process would be quite labor-intensive due to the high density of objects present in the images. The sheer volume of objects necessitated a significant amount of time and effort if we were to rely solely on manual annotation methods, which could lead to delays in our project timeline.

To address this challenge, we opted for a semi-automatic annotation approach utilizing the Segment Anything Model (\textit{SAM}) \cite{kirillov}. This innovative model allows for a more efficient annotation process by leveraging machine learning techniques to assist in the identification and classification of objects within the images. By employing \textit{SAM}, we were able to significantly reduce the time required for annotation while maintaining a high level of accuracy.

At the same time, this model tended to underestimate the boundaries of the pellets, so the annotations had to be subjected to morphological operations. In contrast to classical dilation, we use expanding labels, which do not let connected components expand into neighboring connected components with lower label numbers.

In terms of classification, we established four distinct categories to facilitate the annotation process: ``good'', ``joint'', ``ugly'', and ``big''. These classifications were designed to capture the various characteristics of the objects we were analyzing, allowing for a nuanced understanding of their features and relationships.

Through the use of \textit{SAM}, we successfully completed approximately 80 \%   of the annotations automatically. The remaining 20 \%  of the annotations were performed manually to ensure that any nuances or specific details that the model might have missed were accurately captured. This combination of automated and manual efforts enabled us to efficiently compile a comprehensive dataset for training our model.

Ultimately, this semi-automatic annotation strategy not only expedited the data collection process but also ensured that we had a robust and well-annotated dataset, which is crucial for the subsequent training and performance of our model. By balancing automation with manual oversight, we were able to achieve a high-quality dataset in a timely manner, setting a solid foundation for the success of our project (Table 1).

\begin{table}[!htbp]
\renewcommand{\arraystretch}{1.3}
\centering
\begin{adjustbox}{max width=\textwidth}
\begin{tabular}{p{2.82cm}p{2.84cm}p{3.08cm}p{4.11cm}}
\hline
\multicolumn{1}{|p{2.82cm}}{\begin{center}
		``nice''
\end{center}} & 
\multicolumn{1}{|p{2.84cm}}{\begin{center}
		``ugly''
\end{center}} & 
\multicolumn{1}{|p{3.08cm}}{\begin{center}
		``big''
\end{center}} & 
\multicolumn{1}{|p{4.11cm}|}{\begin{center}
		``joint''
\end{center}} \\ 
\hline
\multicolumn{1}{|p{2.82cm}}{\textit{round, without overlaps or inclusions,} \newline
\textit{no adhering elements, small shape with average distribution} \newline
\textit{the sizes of good pellets are approximately equal and easily visually distinguishable.)} \newline
\textit{green on annotations}} & 
\multicolumn{1}{|p{2.84cm}}{\textit{often either oval or rod-shaped, or of a different shape,} \newline
\textit{unlike most, it has overlaps on the surface.} \newline
\textit{red on annotations}} & 
\multicolumn{1}{|p{3.08cm}}{\textit{large stones of regular round shape, similar to the nicely shaped ones, but larger. } \newline
\textit{unlimited by the upper limit of the size criterion, the size is larger than a regular pellet without sticking.} \newline
\textit{purple on annotations}} & 
\multicolumn{1}{|p{4.11cm}|}{\textit{irregularly shaped pellets are formed by the adhesion of smaller pellets to each other, forming a structure resembling stone.} \newline
\textit{blue on annotations}} \\ 
\hline
\multicolumn{1}{|p{2.82cm}}{\centering
\includegraphics[width=2.67cm,height=2.48cm]{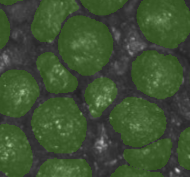}
} & 
\multicolumn{1}{|p{2.84cm}}{\centering
\includegraphics[width=2.69cm,height=2.44cm]{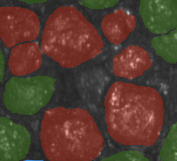}
} & 
\multicolumn{1}{|p{3.08cm}}{\centering
\includegraphics[width=3.01cm,height=2.45cm]{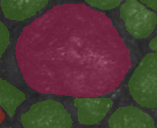}
} & 
\multicolumn{1}{|p{4.11cm}|}{\centering
\includegraphics[width=4.35cm,height=2.5cm]{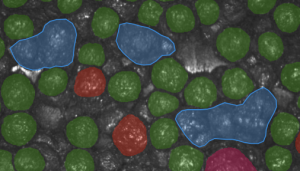}
} \\ 
\hline
\end{tabular}
\end{adjustbox}
\end{table}
\begin{center}
	{\scriptsize Table 1.Description of classes.}
\end{center}

At the next stage, we needed to create a division into train-test subsets. Classic instance-based approaches do not allow for a sufficiently stratified sample—class sizes can vary quite widely. 

Therefore, we implemented a method based on the Wasserstein distance \cite{victor2018} to estimate the distributions of pixels in the subsets. This allowed us to split the pixel-sensitive dataset: when splitting by objects, we received unstratified splits by pixels.

\hfill \break
\begin{equation}
\mathbf{\int dx }\mathbf{\parallel x- }\mathbf{ T }_{\mathbf{p}\xrightarrow[\mathbf{ }]{}\mathbf{q }}\left(\mathbf{x}\right)\mathbf{\parallel }^{\mathbf{2}}\mathbf{p}\left(\mathbf{x}\right)\mathbf{: }\mathbf{L}^{\mathbf{2}}\mathbf{-Wasserstein\  distance}
\end{equation}
\hfill \break

The statement refers to a method of analyzing a dataset by focusing on the distribution of pixel classes rather than individual instances. This approach allows for a more nuanced understanding of the dataset's composition, particularly in contexts such as image processing or computer vision.

By evaluating fractions of the dataset based on pixel distribution, researchers can gain insights into how different classes (such as objects or features within images) are represented across the entire dataset. This is particularly useful in scenarios where the number of instances may not accurately reflect the prevalence or significance of certain classes. For example, in a dataset containing images of various objects, one class may have a large number of instances but occupy a small portion of the total pixel area, while another class may have fewer instances but cover a larger pixel area.

This pixel-based evaluation can lead to more informed decisions regarding model training, as it highlights the importance of class representation in terms of visual content rather than just the count of instances. It can also help in identifying potential biases in the dataset, ensuring that models are trained on a balanced representation of classes, which is crucial for achieving accurate and fair outcomes in machine learning applications.

\subsection{Training}

The backbone chosen was a pretrained EfficientNetV2-b0 \cite{tan} (for low latency and throughput purposes), used from the timm library (Fig. 3).  

\begin{figure}[!htbp]
\centering
\includegraphics[width=9.41cm,height=9.95cm]{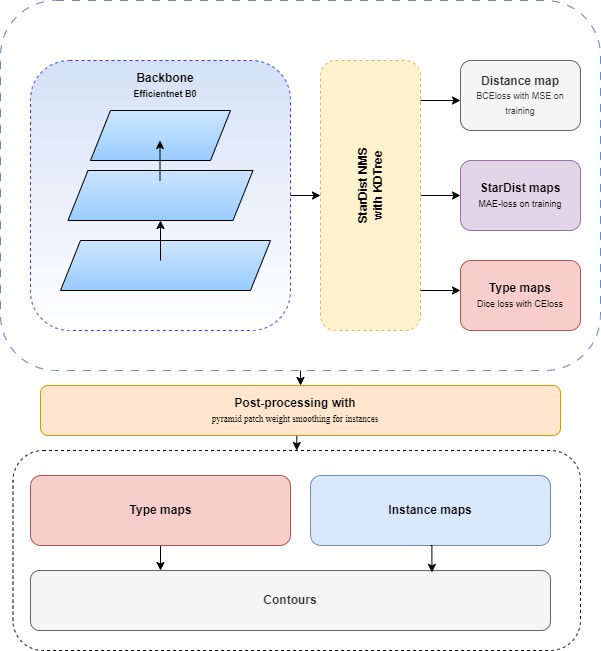}
\end{figure}
\begin{center}
	{\scriptsize Fig. 3. Model diagram.}
\end{center}

The algorithm assumes the return of several types of outputs: distance, type, and stardist. Each branch is trained with a separate loss function, and the final loss, which is used for the optimizer step, is the sum of the losses of all branches of the algorithm.

The distance maps are used as the object probability to determine whether the model detects objects as pellets or not. The stardist maps are used as the input for the \textit{NMS}-based Stardist post-processing pipeline, which is used to separate clumped and overlapping pellets from each other to produce an instance map. We use 32 rays for each object in our task. The type maps are used to classify the pellet instances into predefined type classes (Fig. 4). \cite{gangal}

\begin{figure}[!htbp]
\centering
\includegraphics[width=12.18cm,height=4.16cm]{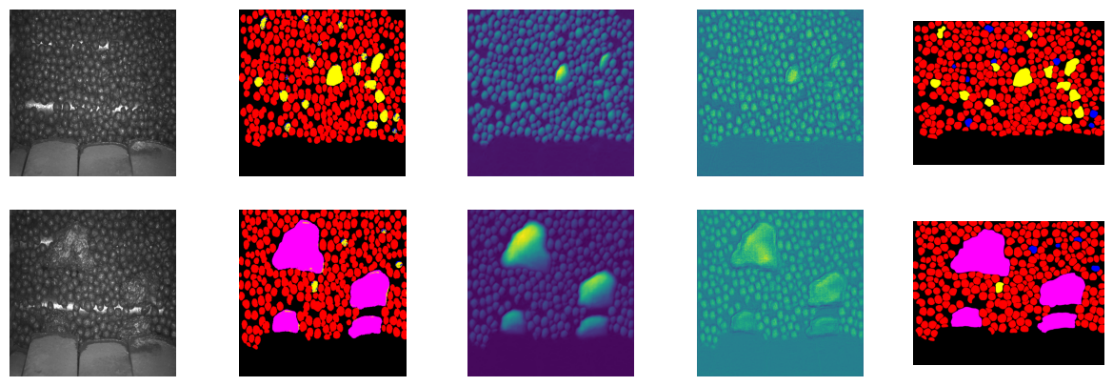}
\end{figure}
\begin{center}
	{\scriptsize Fig. 4. Different masks-output from algorithm.}
\end{center}

For the ``stardist'' output, we set the masked \textit{MAE}. For the ``distance'' outputs, we use the masked \textit{BCELoss} combined with \textit{MSE}, and for the ``type'' output --- the masked multiclass categorical \textit{CELoss} combined with \textit{DiceLoss}. The following weights were used for the losses: distance: 1.0, type: 1.0, stardist: 0.5. These coefficients provided more attention to classes and distances to objects.

\hfill \break
\hfill \break
\begin{equation}
\mathbf{Los}\mathbf{s}_{\mathbf{final}}\mathbf{ = dis}\mathbf{t}_{\mathbf{loss}}\mathbf{\ast 1.0+typ}\mathbf{e}_{\mathbf{loss}}\mathbf{\ast 1.0+stardis}\mathbf{t}_{\mathbf{loss}}\mathbf{\ast 0.5}
\end{equation}
\hfill \break

AdamW with a low learning rate of 3 $\ast$ e$^{-4}$ was chosen as the optimizer for head and backbone, and weight decay with a value of 1 $\ast$ e$^{-5}$ . The model was trained using AMP on half-precision. The classic MultiStepLR scheduler was used with steps at 500 and 800 epochs out of 1000 in total.

Morphological transformations were applied only to gt-masks, since they were mostly generated using auto-labeling. Blur and hue saturation were used as augmentations. The training image size is 1000 by 1000 pixels, not compressed but cropped in the center to avoid interpolation effects on objects. To speed up post-processing, JIT compilation methods for functions were used.

To determine the contours after training, we generated a binary mask of the detected objects. Then, we iterated through all the objects and copied the region on images within the object areas. After that, we applied classical computer vision methods to find the contours.

To find dimensions, we used a bounding circle for each object, filtering out those formed by fewer than 8 points. The diameter of such a circle was taken as the final pellet size.

\subsection{Metrics}

The classic \textit{IoU}-based metric from segmentation models pytorch has been adopted as a metric \cite{kirillov2019panoptic}: a detected object $I_{\text{pred}}$ is considered a match (true positive \textit{TP}) if there is an underlying true object $I_{\text{gt}}$ whose intersection over union (\textit{IoU}) is greater than a given threshold $\tau \in [0, 1]$. Predicted objects that do not match are considered false positives (\textit{FP}). The metric was calculated both for each class and for the entire dataset and averaged over the epoch. Additionally, the classic precision and recall were used for each class for comparison with other methods \cite{wasserman2018topological} \cite{soille}.

\subsection{Additional Methods}

The methods employed in this study aimed to develop a classifier based on Vision Transformer (ViT) for distinguishing between two classes. However, the results were not satisfactory; the classifier exhibited significant instability and lacked sufficient interpretability, making it difficult to understand its decision-making process.

To address these shortcomings, an alternative approach was proposed involving a technique known as cross marking. This method entails combining six images into a single composite image and then soliciting input from multiple individuals to identify the ``dirtiest'' image that contains defective pellets. By aggregating the ratings from different evaluators, we can derive a probability score for each image's classification, which opens the door to employing a regression-based approach for more nuanced analysis \cite{jaderberg2015spatial}.

Additionally, the search for defective pellets could be enhanced through the application of anomaly detection algorithms \cite{yufastflow}. While the current model demonstrates some responsiveness to specific areas of larger objects, it also struggles with translucent sections of rollers, leading to a high rate of missed detections. Consequently, the model's performance is insufficient for productive use in its current form, as it frequently fails to identify actual defects. Further refinement and optimization of the model are necessary to improve its reliability and effectiveness in detecting bad pellets (Fig. 5).

\begin{figure}[!htbp]
\centering
\includegraphics[width=8.42cm,height=4.27cm]{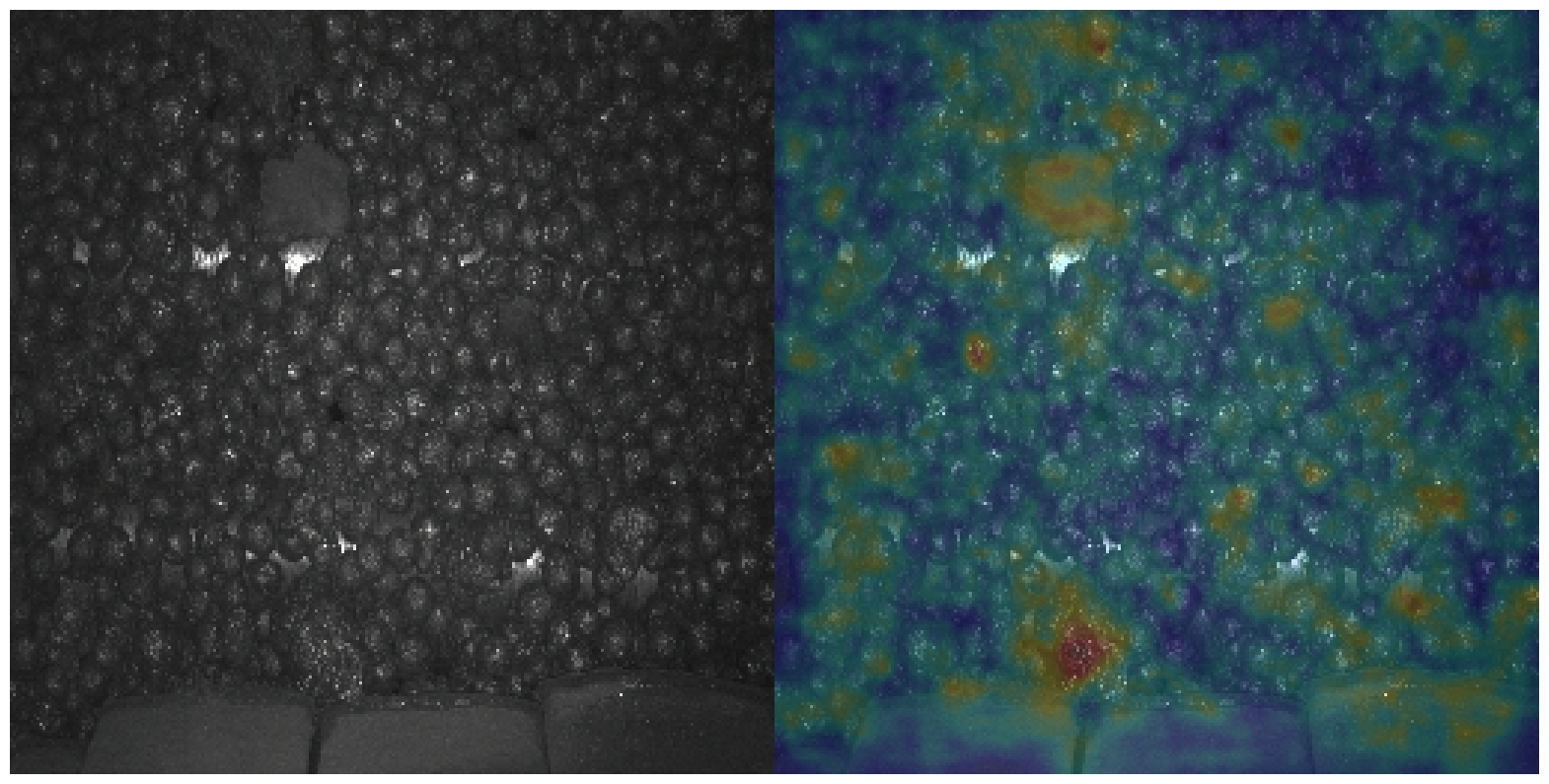}

\begin{center}
	{\scriptsize Fig. 5. FastFlow.}
\end{center}
\end{figure}

The interpretation of such masks is difficult and does not allow solving the problem with one model. Therefore, we explored additional approaches.

A metric learning approach for classification was carried out by dividing frames into two classes: \textit{ugly} and \textit{nice}. This approach is similar to a \textit{VIT}-based classifier, but with the addition of another loss function \cite{lin2017focal}. However, this approach has disadvantages: we examine the entire frame—false positives are possible, and it is difficult to attribute a frame to one class or the other.

We applied a pre-trained \textit{SAM} model (Fig. 6) to photos of pellets, highlighting the contours of the pellets and attempting to identify ``anomalous'' contours in the photos using \textit{IsolationForest} to confirm the hypothesis that the ugly pellets would be anomalies in the images. However, this approach did not allow us to isolate the number of defective pellets: we could see areas that deviated from the norm, but we could not say with certainty how much the production process had deviated from the specified standards.

\begin{figure}[!htbp]
\centering
\begin{subfigure}[b]{0.3\textwidth}
\centering
\includegraphics[width=\textwidth]{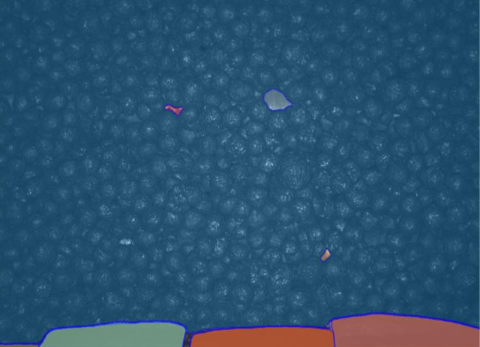}
\end{subfigure}
\hfill
\begin{subfigure}[b]{0.3\textwidth}
\centering
\includegraphics[width=\textwidth]{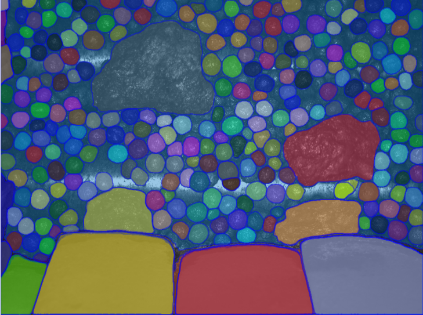}
\end{subfigure}
\hfill
\begin{subfigure}[b]{0.3\textwidth}
\centering
\includegraphics[width=\textwidth]{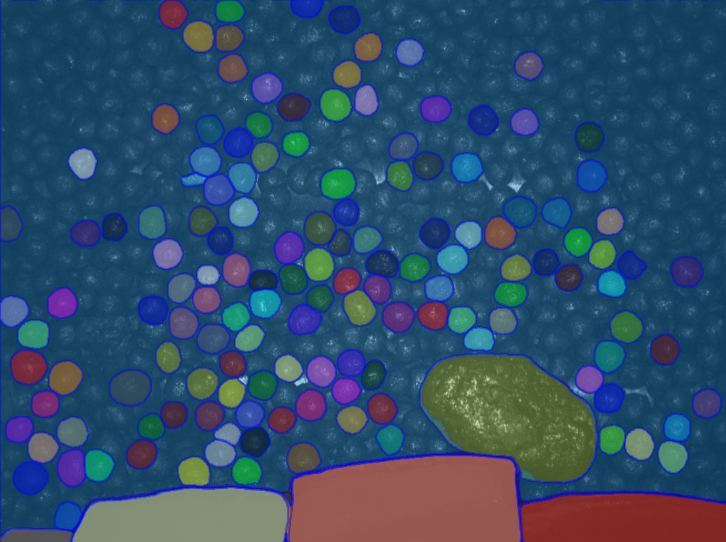}
\end{subfigure}
\end{figure}
\begin{center}
	{\scriptsize Fig. 6. Instance segmentation masks by Segment Anything method w/o tuning.}
\end{center}

This model generally identifies masks of well-demarcated objects, but it can make mistakes and divide one large object into several or, conversely, combine several small ones into one mask—thus yielding non-robust and non-consistent results.
The most comparable results were obtained using Mask R-CNN \cite{he2018maskrcnn}, so further comparisons were made with this model. The problem with the detector in this case was that it first identifies bounding boxes and then segments the objects. There was also significant class confusion. However, the algorithm performed comparably at first glance (Fig. 7).
 
\pagebreak
\begin{figure}[!htbp]
\centering
\includegraphics[width=12.84cm,height=3.2cm]{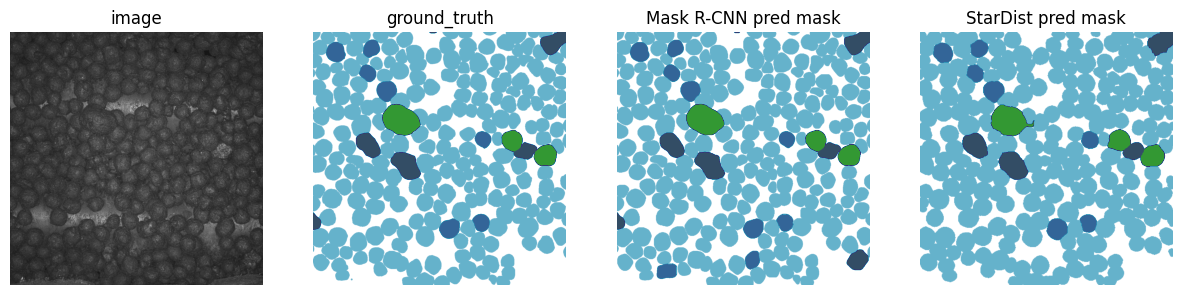}
\end{figure}
\begin{figure}[!htbp]
\centering
\includegraphics[width=12.84cm,height=3.2cm]{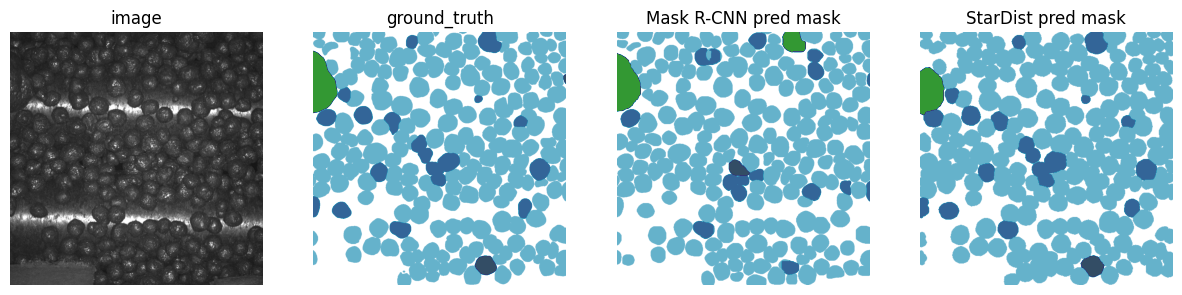}
\end{figure}
\begin{figure}[!htbp]
\centering
\includegraphics[width=12.84cm,height=3.2cm]{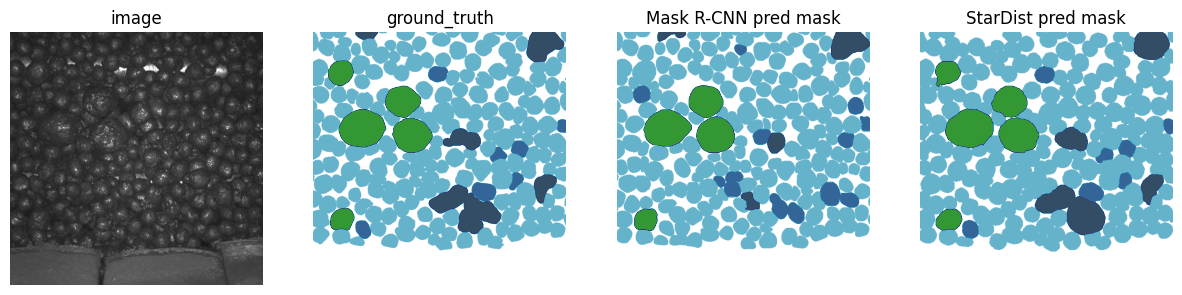}
\end{figure}
\begin{center}
	{\scriptsize Fig. 7. Stardist detections vs maskRCNN. The colors reflect the different classes.}
\end{center}

Also, the problem with the detector was that it tended to detect all the pellets that came across the frame, while the accuracy of their identification (classification) was important to us. 
And of course, we tried the classic segmenter U-Net++ \cite{unet}, but despite good metrics, we encountered issues with fused masks that were not observed with the detector or confused classes (Fig. 8). However, one way or another, the model was included in the comparison.

\begin{figure}[!htbp]
	\centering
\includegraphics[width=12.84cm,height=3.2cm]{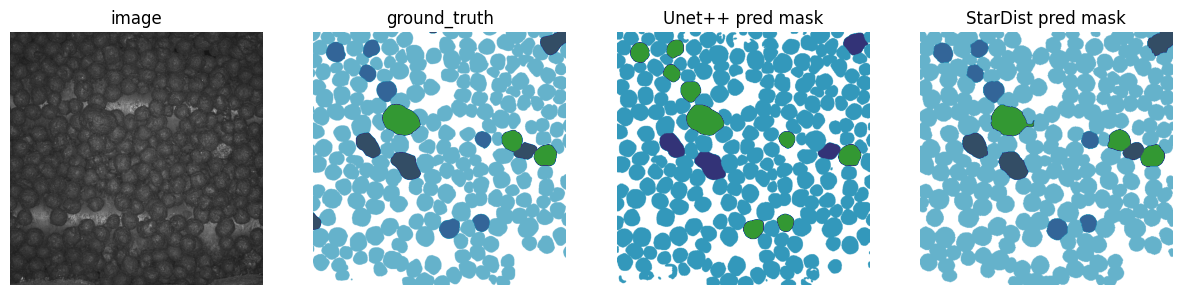}
\end{figure}
\begin{figure}[!htbp]
	\centering
\includegraphics[width=12.84cm,height=3.2cm]{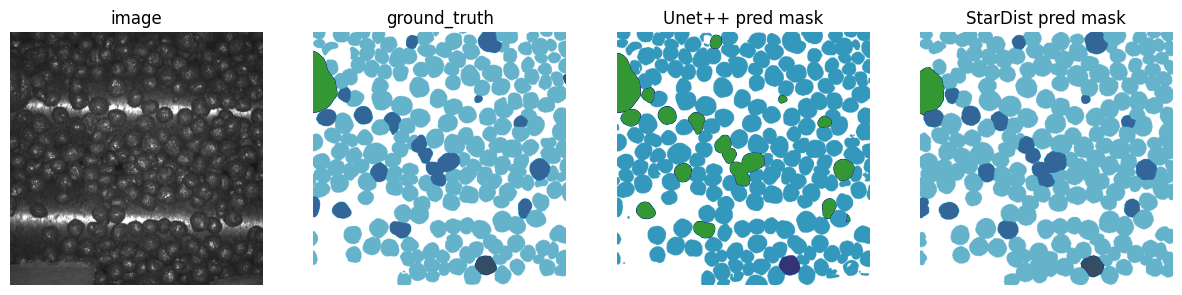}
\end{figure}

\begin{center}
	{\scriptsize Fig. 8. Unet++ preds. The colors reflect the different classes.}
\end{center}
\hfill
\break
\begin{figure}[!htbp]
	\centering
\includegraphics[width=12.84cm,height=3.2cm]{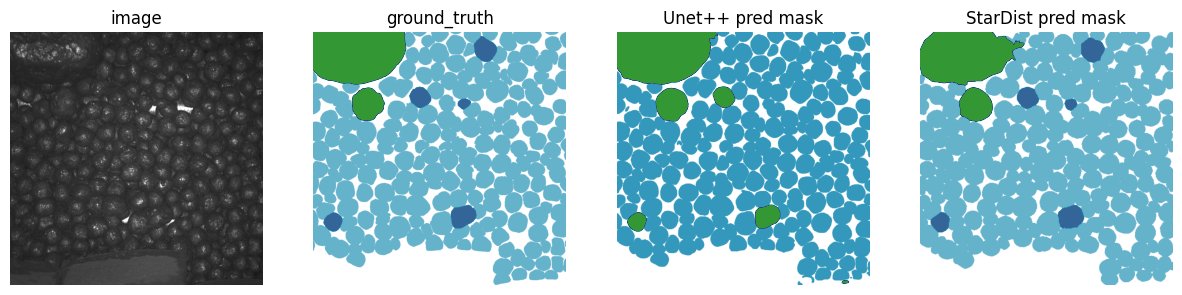}
\end{figure}
\begin{center}
	{\scriptsize Fig. 8. Unet++ preds. The colors reflect the different classes (continue).}
\end{center}

\section{Results}

We achieved highly accurate results in the segmentation of pellets, enabling us to effectively determine class distributions within the images analyzed. This methodology facilitates the precise separation of objects, ensuring a high level of accuracy in our assessments.

Our testing was conducted on a deferred dataset, which yielded impressive metrics: an accuracy of \textit{0.991564} and an Intersection over Union (IoU) score of \textit{0.961676}. Additionally, we recorded a precision and recall rate of \textit{0.98039} and \textit{0.98021}. These outstanding results empower us to accurately distinguish between defective pellets and those that meet quality standards, a critical capability for optimizing production processes. By implementing this advanced segmentation technique, we can enhance quality control measures and improve overall operational efficiency in the manufacturing environment (Fig.~9).

\begin{table}[!htbp]
\centering
\renewcommand{\arraystretch}{1.3}
\begin{adjustbox}{max width=\textwidth}
\begin{tabular}{p{6.19cm}p{6.15cm}}
\multicolumn{1}{p{6.41cm}}{\includegraphics[width=5.39cm,height=4.04cm]{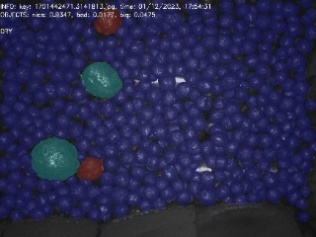}
} & 
\multicolumn{1}{p{6.41cm}}{\includegraphics[width=5.39cm,height=4.04cm]{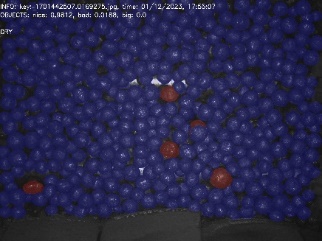}
} \\ 
\multicolumn{1}{p{6.41cm}}{\includegraphics[width=5.39cm,height=4.04cm]{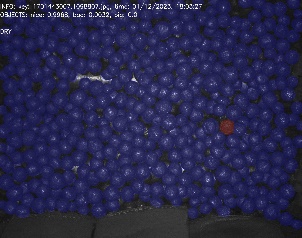}
} & 
\multicolumn{1}{p{6.41cm}}{\includegraphics[width=5.39cm,height=4.04cm]{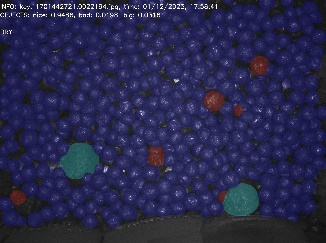}
} \\ 
\end{tabular}
\end{adjustbox}
\end{table}
\begin{center}
	{\scriptsize Fig. 9. Instance segmentation masks on images by StarDists.}
\end{center}

Furthermore, the exceptional accuracy achieved in the segmentation process enables the precise identification of the pellets' contours. This capability allows for the conversion of the segmented dimensions into their corresponding physical measurements in the real world. The reliability of this method has been further substantiated through rigorous laboratory experiments, which have confirmed the accuracy of the measurements obtained. These experiments serve to validate the effectiveness of the segmentation technique, ensuring that the dimensions derived from the digital analysis align closely with actual physical properties. This level of precision is crucial for applications that require exact measurements, thereby enhancing the overall utility and applicability of the segmentation process in various fields (Fig.~10).

\begin{table}[!htbp]
\centering
\renewcommand{\arraystretch}{1.3}
\begin{adjustbox}{max width=\textwidth}
\begin{tabular}{p{6.19cm}p{6.15cm}}
	\multicolumn{1}{p{6.41cm}}{\includegraphics[width=5.48cm,height=4.11cm]{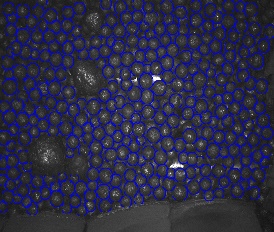}
	} & 
	\multicolumn{1}{p{6.41cm}}{\includegraphics[width=5.39cm,height=4.04cm]{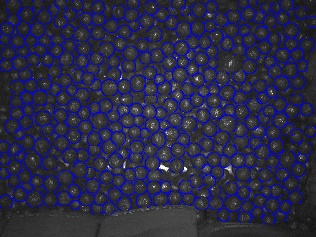}
	} \\ 
	\multicolumn{1}{p{6.41cm}}{\includegraphics[width=5.48cm,height=4.19cm]{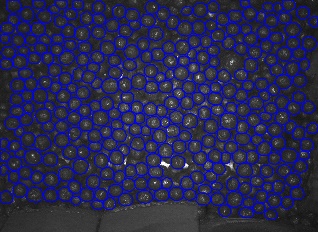}
	} & 
	\multicolumn{1}{p{6.41cm}}{\includegraphics[width=5.42cm,height=4.11cm]{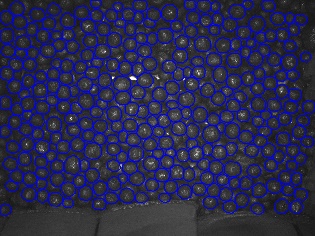}
	} \\ 
\end{tabular}
\end{adjustbox}

\begin{center}
	{\scriptsize Fig. 10. Contours of pellets.}
\end{center}
\end{table}

Several approaches were used for contours: circle, convex hull, ellipse. The approach using circles proved to be the best \cite{kirillovinstancecut}. Measurements are made only on ``nice''. pellets, so this form conveyed the physical dimensions of the object more accurately. For your tasks, it is recommended to rely on the shape of the measured object.

Below are the metrics in comparison with another model in terms of determining belonging to certain types of pellets. To assess how dimensions are measured, a different approach was used. Additionally, we provide a comparison of training on data prepared with and without Wasserstein distance, as well as other approaches (Table 2).

\hfill \break
\hfill \break
\begin{table}[!htbp]
\centering
\renewcommand{\arraystretch}{1.3}
\begin{adjustbox}{max width=\textwidth}
\begin{tabular}{p{4.21cm}p{2.87cm}p{3.25cm}p{2.51cm}}
\hhline{~~~~}
\multicolumn{1}{p{4.21cm}}{} & 
\multicolumn{1}{p{2.87cm}}{\textbf{IoU of masks}} & 
\multicolumn{1}{p{3.25cm}}{\textbf{Precision}} & 
\multicolumn{1}{p{2.51cm}}{\textbf{Recall}} \\ 
\hline
\multicolumn{1}{p{4.21cm}}{MaskRCNN} & 
\multicolumn{1}{p{2.87cm}}{0.933487} & 
\multicolumn{1}{p{3.25cm}}{0.95049} & 
\multicolumn{1}{p{2.51cm}}{0.96528} \\ 
\hline
\multicolumn{1}{p{4.21cm}}{Unet++} & 
\multicolumn{1}{p{2.87cm}}{0.949782} & 
\multicolumn{1}{p{3.25cm}}{0.94457} & 
\multicolumn{1}{p{2.51cm}}{0.97098} \\ 
\hline
\multicolumn{1}{p{4.21cm}}{StarDist-based EfficientNet b0 with Wasserstein data (best)} & 
\multicolumn{1}{p{2.87cm}}{\textbf{0.961676}} & 
\multicolumn{1}{p{3.25cm}}{\textbf{0.98039}} & 
\multicolumn{1}{p{2.51cm}}{0.98021} \\ 
\hline
\multicolumn{1}{p{4.21cm}}{StarDist-based EfficientNet b0 w/o Wasserstein data} & 
\multicolumn{1}{p{2.87cm}}{0.934291} & 
\multicolumn{1}{p{3.25cm}}{0.97327} & 
\multicolumn{1}{p{2.51cm}}{0.97904} \\ 
\hline
\multicolumn{1}{p{4.21cm}}{StarDist-based EfficientNet b1 w Wasserstein data} & 
\multicolumn{1}{p{2.87cm}}{0.960269} & 
\multicolumn{1}{p{3.25cm}}{0.97983} & 
\multicolumn{1}{p{2.51cm}}{\textbf{0.98327}} \\ 
\hhline{~~~~}
\end{tabular}
\end{adjustbox}

\begin{center}
	{\scriptsize Table 2. Metrcis summary.}
\end{center}
\end{table}
\hfill
\break
\hfill \break
\hfill \break
\pagebreak
\begin{figure}[!htbp]
\centering
\includegraphics[width=12.84cm,height=3.09cm]{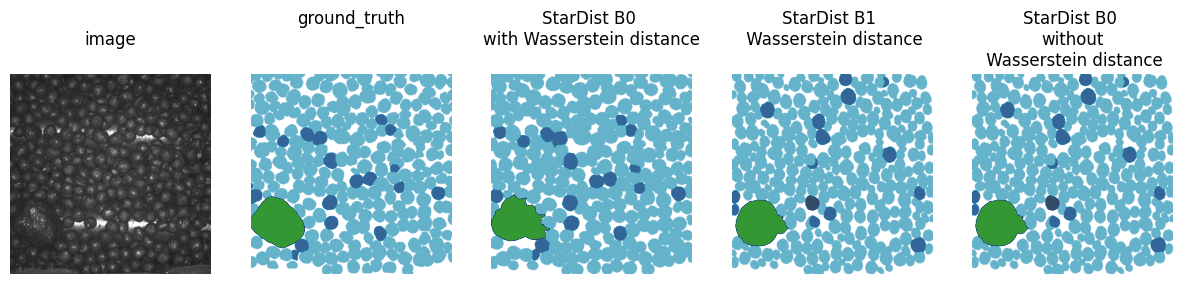}
\end{figure}

\begin{figure}[!htbp]
\centering
\includegraphics[width=12.84cm,height=3.09cm]{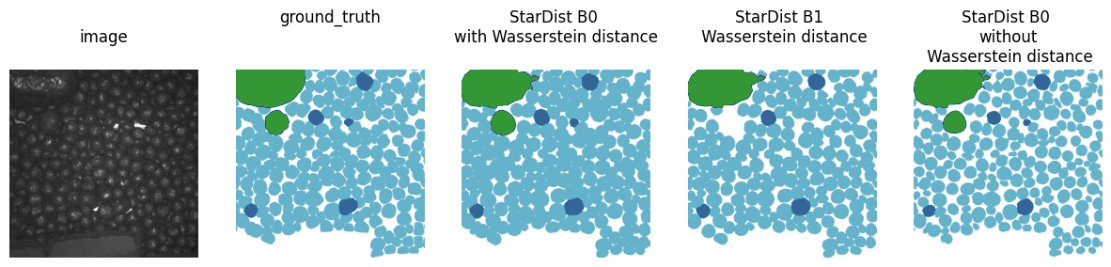}
\end{figure}

\begin{figure}[!htbp]
\centering
\includegraphics[width=12.84cm,height=3.09cm]{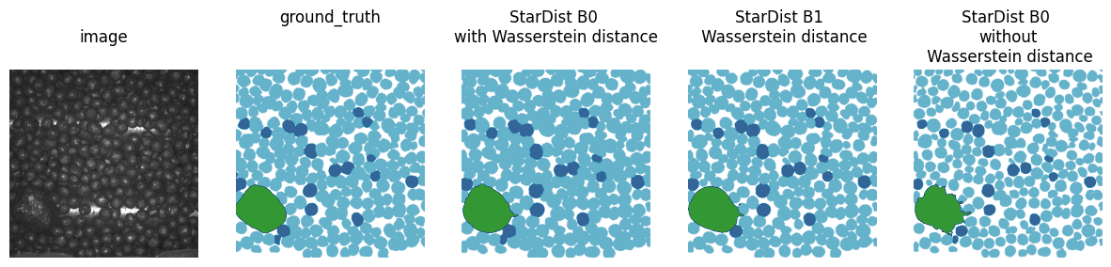}

\begin{center}
	\centering
	{\scriptsize Fig. 11. Visual comprasion of different approaches: 
		b0 w/o wassertein, b1 with wasserstein, b0 with wasserstein. The colors reflect the different classes.}
\end{center}
\end{figure}

\section{Validation}

To ensure the accuracy and reliability of our findings, we conducted a series of laboratory experiments focused on measuring the dimensions and characteristics of pellets over a predetermined fixed period. This approach allowed us to gather empirical data that could be systematically analyzed~\cite{soille}.

In our methodology, we utilized model outputs to establish a benchmark for identifying the size and quality of what we classified as ``good pellets.'' These model outputs served as a theoretical framework, guiding our expectations and criteria for pellet quality~\cite{xiaoyanimage}.

Subsequently, we performed a comparative analysis between the model-derived specifications and the actual measurements obtained from our laboratory studies. This comparison was crucial in validating our model's predictions and ensuring that our theoretical assumptions aligned with practical observations. By correlating the model outputs with the experimental data, we aimed to enhance the robustness of our conclusions and provide a comprehensive understanding of pellet quality metrics.

Laboratory experiments were conducted over a period of 2 months at various times. On average, we achieved a Mean Absolute Error (MAE) of $\sim$ 4.9 $\pm$ 0.1. Statistical hypotheses were not tested because the laboratory samples were collected from a specially designated location rather than directly under the camera. Our focus was on the selection time, and this approach enabled us to assess the tendencies and trends of product quality, which is sufficient for analyzing the production process. This analysis can be utilized by machine learning algorithms or mathematical models (Fig. 12).
\pagebreak
\begin{figure}[!htbp]
	\centering
	\includegraphics[width=12.84cm,height=5.42cm]{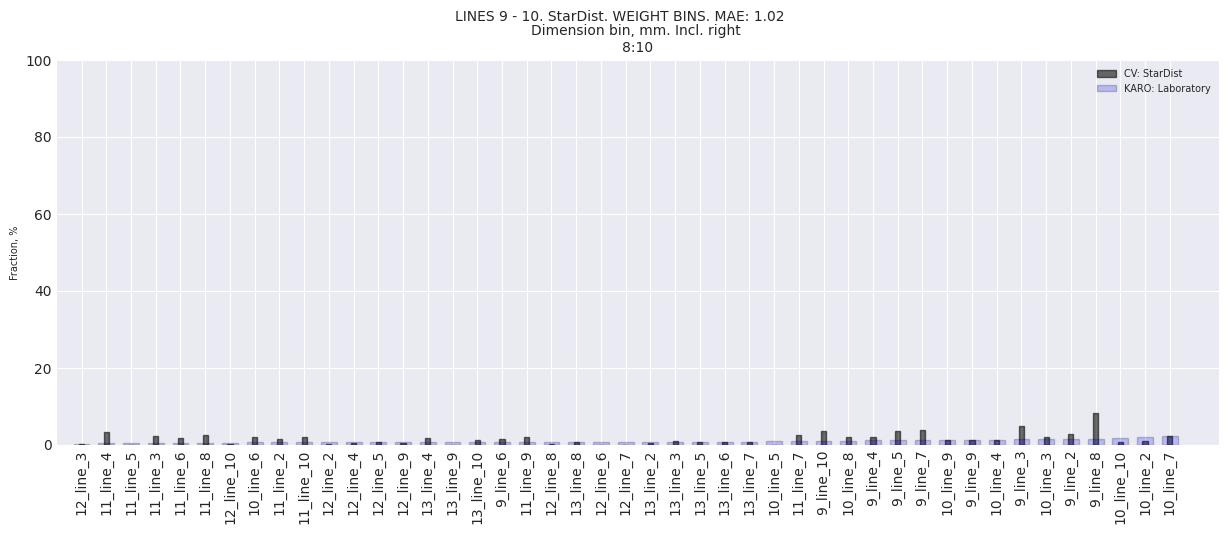}
\end{figure}
\begin{figure}[!htbp]
	\centering
	\includegraphics[width=12.84cm,height=5.42cm]{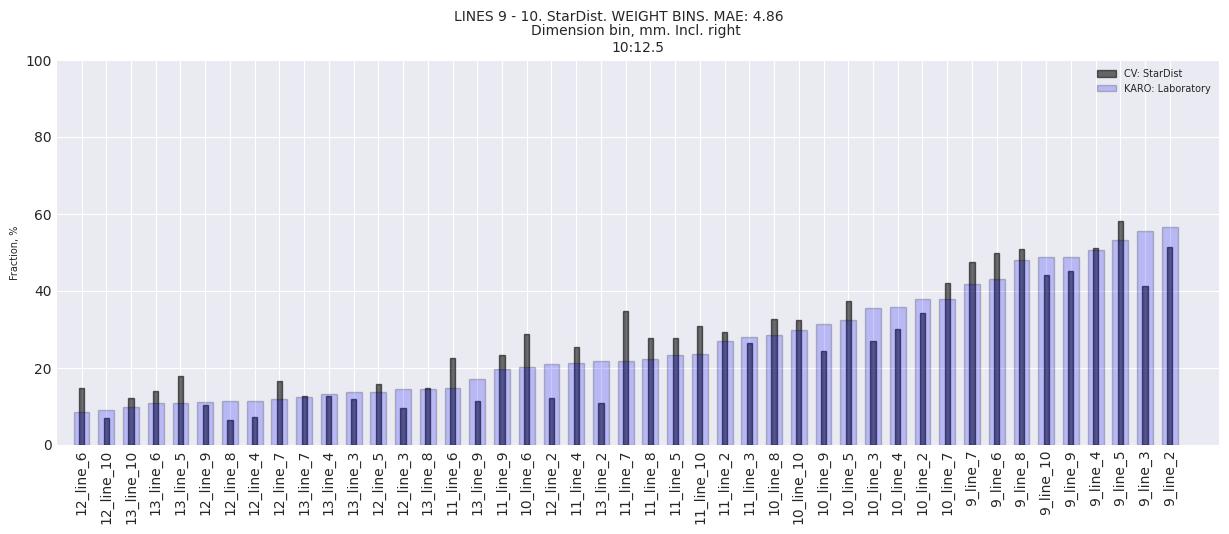}
\end{figure}
\begin{figure}[!htbp]
	\centering
	\includegraphics[width=12.84cm,height=5.42cm]{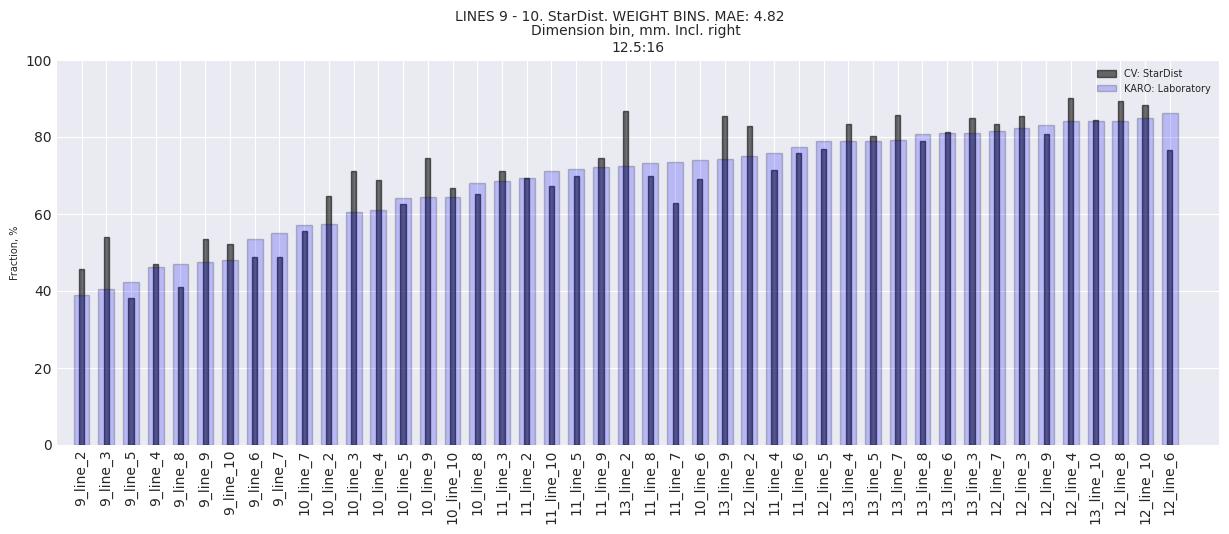}
\begin{center}
	{\scriptsize Fig. 12. Comparison with laboratory experiments. Black is our model, blue-laboratory.}
\end{center}
\end{figure}

\section{Problems}

The research revealed multiple issues within the algorithm that significantly impact the overall performance of the system, particularly in the areas of contour detection and classification accuracy (Fig.~13).

One critical observation is the importance of monitoring the brightness levels captured by the cameras. Our experiments involved systematically altering the luminance settings of the images used in the analysis. We found that variations in brightness had a detrimental effect on the quality of the output. Specifically, as the luminance was adjusted, there were instances where the algorithm's ability to make accurate predictions diminished considerably. In some cases, the changes in brightness were so severe that they resulted in a complete failure to generate any predictions at all.

This underscores the necessity for careful calibration and control of lighting conditions during data capture to ensure optimal performance of the algorithm. Addressing these brightness-related issues is essential for enhancing the reliability and effectiveness of the system in real-world applications.

\begin{figure}[!htbp]
\begin{subfigure}[b]{0.45\textwidth}
\includegraphics[width=\textwidth]{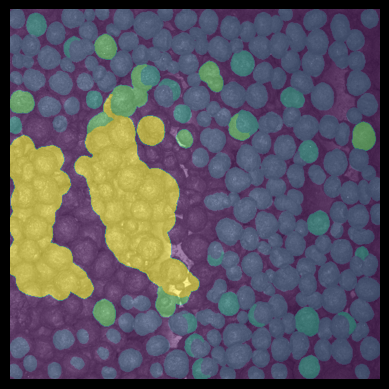}
\end{subfigure}
\hfill
\begin{subfigure}[b]{0.45\textwidth}
\includegraphics[width=\textwidth]{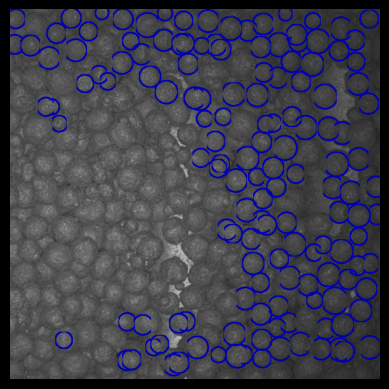}
\end{subfigure}

\begin{center}
	{\scriptsize Fig. 13. Corrupted detections.}
\end{center}
\end{figure}

The statement refers to a process of normalizing the luminance channel of an image to align it with a specific standard within a dataset, specifically using the CIELAB color space.

In this context, the luminance channel represents the brightness of the image, which can significantly influence the performance of a model, particularly in tasks related to image classification or segmentation. By adjusting the luminance channel to a general distribution, you are effectively standardizing the brightness levels across the dataset. This normalization helps to mitigate variations in lighting conditions or exposure that could otherwise skew the model's performance.

By normalizing the luminance channel, the model's ability to accurately identify and delineate the contours of objects can be enhanced, leading to improved performance metrics.

In summary, this process of luminance normalization is crucial for ensuring that the model can generalize better across different images by reducing the impact of lighting variations, thereby allowing for more consistent and reliable performance in identifying specific classes within the dataset.

To enhance the performance and reliability of the model, we implemented several strategies aimed at normalizing its operations. One key approach involved incorporating hard augmentations specifically targeting brightness variations. These augmentations are designed to artificially increase the diversity of the training data by introducing significant changes in brightness levels, thereby helping the model become more robust to different lighting conditions.

In addition to these augmentations, we also integrated a preprocessing step that adjusts the luminance of the input data. This preprocessing technique is crucial as it aligns the luminance distribution of the input samples with that of the training dataset. By ensuring that the luminance characteristics of the incoming data closely match those encountered during training, we can improve the model's ability to generalize and perform effectively across various scenarios. This dual approach of applying hard augmentations alongside a targeted preprocessing strategy is essential for optimizing the model's performance and ensuring its adaptability to real-world applications (Fig. 14).

\pagebreak
\begin{figure}[!htbp]
	\centering
	\begin{subfigure}[b]{0.45\textwidth}
		\centering
		\includegraphics[width=\textwidth]{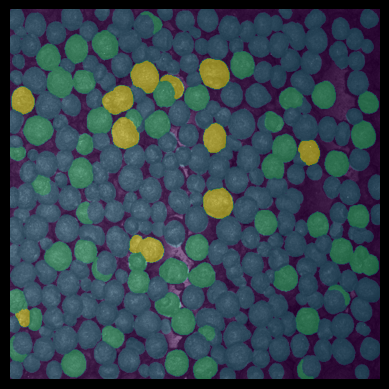}
	\end{subfigure}
	\hfill
	\begin{subfigure}[b]{0.45\textwidth}
		\centering
		\includegraphics[width=\textwidth]{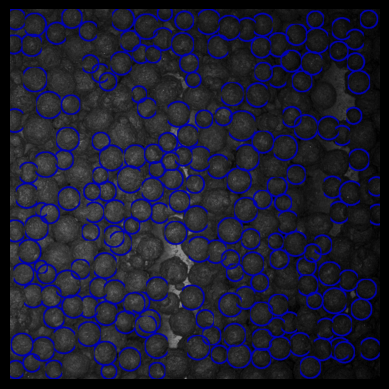}
	\end{subfigure}
\end{figure}
\begin{center}
	{\scriptsize Fig. 14. Fixed detections.}
\end{center}

This approach enhances the model's predictions by increasing their consistency and reliability (Fig.~15). By implementing strategies that focus on reducing variability and improving the accuracy of the outcomes, the model becomes more adept at handling a range of inputs and scenarios. As a result, users can expect more stable and trustworthy predictions, which is particularly important in applications where decision-making is critical. This robustness not only fosters confidence in the model's outputs but also allows for better performance in real-world situations, ultimately leading to more effective and informed decision-making processes.
 
\begin{figure}[!htbp]
\centering
\begin{subfigure}[b]{0.36\textwidth}
\centering
\includegraphics[width=\textwidth]{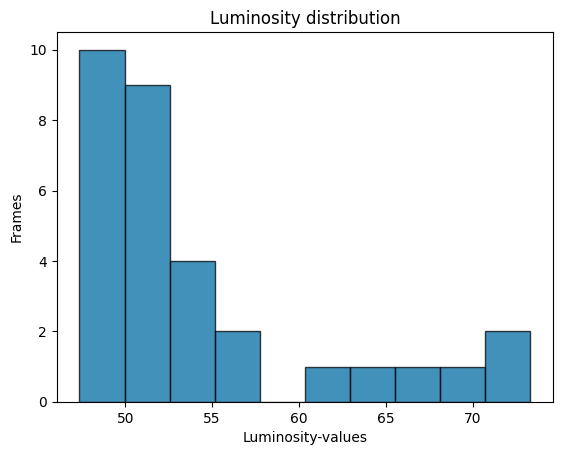}
\end{subfigure}
\begin{subfigure}[b]{0.54\textwidth}
\centering
\includegraphics[width=\textwidth]{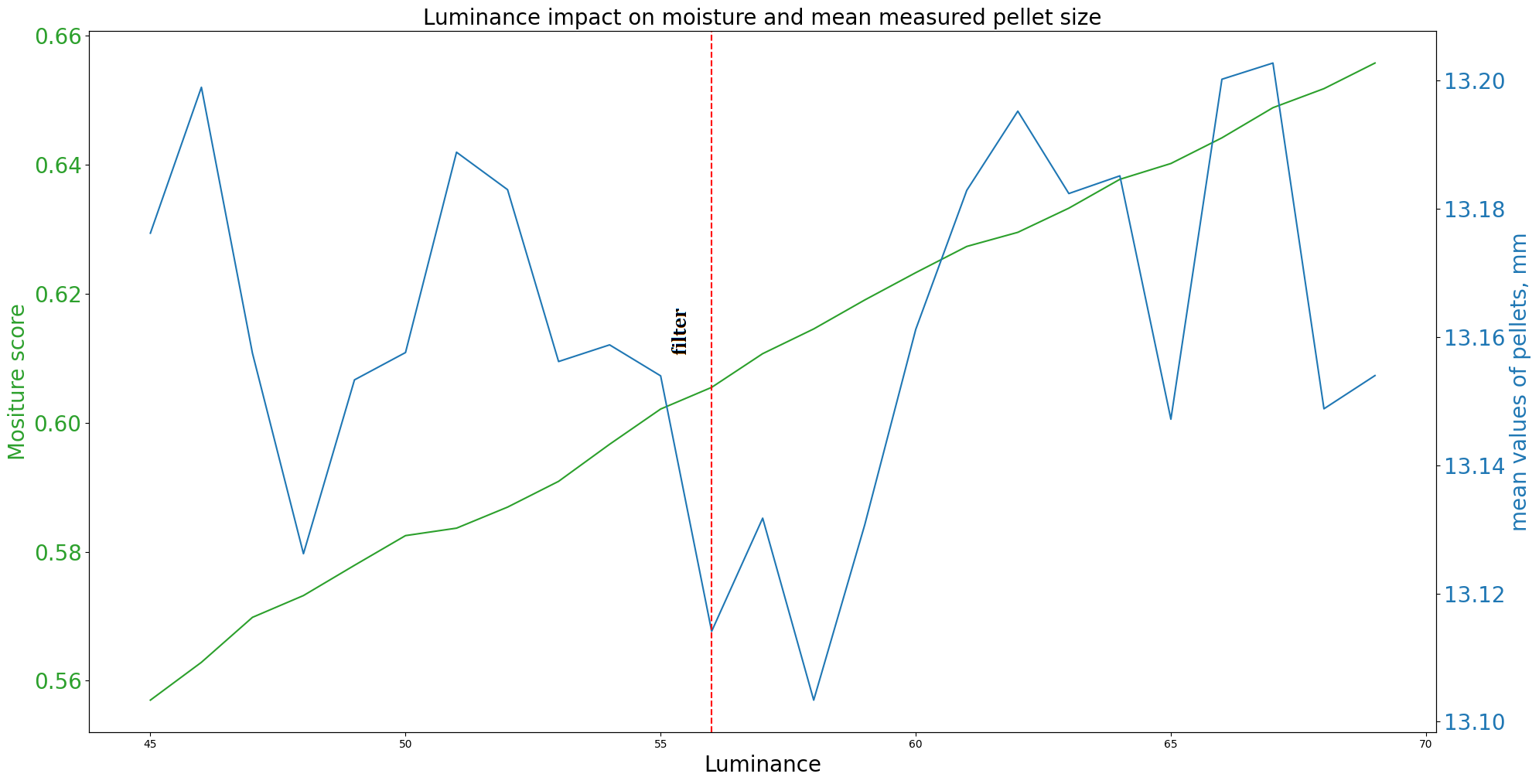}
\end{subfigure}
\end{figure}
\begin{center}
	{\scriptsize Fig. 15. Dependence on luminance.}
\end{center}

\section{Conclusion}

The initial objective of our project was to identify and categorize the various types of pellets. However, as we progressed, the capabilities of our algorithm evolved, allowing us to incorporate additional functionalities, such as analyzing the distribution of pellet sizes. Our research demonstrates that utilizing star polygons serves as an effective geometric representation for the precise localization of iron ore pellets, even in challenging and complex imaging conditions.

This method proves particularly advantageous when working with high-density images, where traditional models may struggle. Unlike more complex frameworks such as Mask R-CNN or U-Net, which often require extensive hyperparameter tuning to optimize performance, our approach with StarDist is streamlined. It features a limited number of hyperparameters, making it easier to achieve satisfactory results without the need for exhaustive adjustments.

Features obtained using precise segmentation can be used to further train additional process analysis algorithms, for example, to determine the humidity of the process or to indicate violations in production technology.

The high accuracy in segmentation provided by our method enables us to accurately determine the physical dimensions of the pellets. This capability is crucial for calculating size distributions, which can have significant implications for various applications in the industry. Furthermore, our approach illustrates how solutions derived from one domain can effectively transfer knowledge and techniques to another, even when the two fields are not directly related. This cross-domain application not only enhances the versatility of our algorithm but also opens up new avenues for research and development.

\hfill
\break
\hfill
\break
\hfill
\break


\begin{thebibliography}{10}

\bibitem{kirillovinstancecut}
Kirillov A, Levinkov E, Andres B, Savchynskyy B, and Rother C.
\newblock Instancecut: from edges to instances with multicut.
\newblock 2016.

\bibitem{kirillov}
Kirillov Alexander, Mintun Eric, Ravi Nikhila, Mao Hanzi, Rolland Chloe,
  Gustafson Laura, Xiao Tete, Whitehead Spencer, Berg Alexander, Lo~Wan-Yen,
  Dollar Piotr, and Girshick Ross.
\newblock Segment anything.
\newblock 2023.

\bibitem{kirillov2019panoptic}
Kirillov Alexander, He~Kaiming, Girshick Ross, Rother Carsten, and Dollar
  Piotr.
\newblock Panoptic segmentation.
\newblock 2019.

\bibitem{gangal}
Gangal Ayushe, Kumar Peeyush, and Kumari Sunita.
\newblock Complete scanning application using opencv.
\newblock 2021.

\bibitem{yufastflow}
Yu~Jiawei, Zheng Ye, Wang Xiang, Li~Wei, Wu~Yushuang, Zhao Rui, and Wu~Liwei.
\newblock Fastflow: Unsupervised anomaly detection and localization via 2d
  normalizing flows.
\newblock 2021.

\bibitem{he2018maskrcnn}
He~Kaiming, Gkioxari Georgia, Dollar Piotr, and Girshick Ross.
\newblock Maskrcnn".
\newblock 2018.

\bibitem{wasserman2018topological}
Wasserman L.
\newblock Topological data analysis.
\newblock 2018.

\bibitem{xiaoyanimage}
Xiaoyan Liu, Chuangang Mao, Sun Wei, and Xin W.
\newblock Image-based method for measuring pellet size distribution in the
  stable area of disc pelletizer.
\newblock 2018.

\bibitem{jaderberg2015spatial}
Jaderberg M, Simonyan K, Zisserman A, and Kavukcuoglu K.
\newblock Spatial transformer networks.
\newblock 2015.

\bibitem{victor2018}
Victor M, Panaretos Yoav, and Zemel.
\newblock Statistical aspects of wasserstein distances.
\newblock 2018.

\bibitem{weigert2022nuclei}
Weigert Martin and Schmidt Uwe.
\newblock Nuclei instance segmentation and classification in histopathology
  images with stardist.
\newblock 2022.

\bibitem{weigert2020star}
Weigert Martin, Schmidt Uwe, Haase Robert, Sugawara Ko, and Myers Gene.
\newblock Star-convex polyhedra for 3d object detection and segmentation in
  microscopy.
\newblock 03 2020.

\bibitem{tan}
Tan Mingxing, Quoc V, and Le.
\newblock Efficientnetv2: Smaller models and faster training.
\newblock 2021.

\bibitem{soille}
Soille P.
\newblock Morphological image analysis: Principles and applications.
\newblock 1999.

\bibitem{lefuvre2007}
Lefuvre S.
\newblock Int. conf. on computer analysis of images and patterns.
\newblock 2007.

\bibitem{graham2021}
Graham Simon, Jahanifar Mostafa, Vu~Quoc, Hadjigeorghiou Giorgos, Leech Thomas,
  and Snead David.
\newblock Conic: Colon nuclei identification and counting challenge 2022.
\newblock 2021.

\bibitem{lin2017focal}
Lin Tsung-Yi, Goyal Priya, Girshick Ross, He~Kaiming, and Dollar Piotr.
\newblock Focal loss for dense object detection.
\newblock 2017.

\bibitem{schmidt2018cell}
Schmidt Uwe, Weigert Martin, Broaddus Coleman, and Myers Gene.
\newblock Cell detection with star-convex polygons.
\newblock 2018.

\bibitem{unet}
Nima Tajbakhsh Jianming~Liang Zongwei~Zhou, Md Mahfuzur Rahman~Siddiquee.
\newblock Unet++: A nested u-net architecture for medical image segmentation.
\newblock 2018.

\end{thebibliography}
\end{document}